# CONFIDENCE FACTORS, EMPIRICISM AND THE DEMPSTER-SHAFER THEORY OF EVIDENCE

by
John F. Lemmer
RCA, Advanced Technology Laboratories
Moorestown, NJ 08057
(609) 866-6650

## ABSTRACT

The issue of confidence factors in Knowledge Based Systems has become increasingly important and Dempster-Shafer (DS) theory has become increasingly popular as a basis for these factors. This paper discusses the need for an empirical interpretation of any theory of confidence factors applied to Knowledge Based Systems and describes an empirical interpretation of DS theory suggesting that the theory has been seriously misinterpreted. For the essentially syntactic DS theory, the empirical model developed is based on the semantics of sample spaces. This model is used to show that, if belief functions are based on reasonably accurate sampling or observation of a sample space, then the beliefs and upper probabilities as computed according to DS theory cannot be interpreted as frequency ratios. Since a number of proposed applications of DS theory use belief functions in situations with statistically derived evidence and seem to appeal to statistical intuition to provide an interpretation of the results, it is likely that DS theory has often been misapplied. Examples are cited.

# CONFIDENCE FACTORS, EMPIRICISM
# AND
# THE DEMPSTER-SHAFER THEORY OF EVIDENCE

The issue of confidence factors in Knowledge Based Systems has become increasingly important and Dempster-Shafer (DS) theory has become increasingly popular as a basis for these factors. This paper discusses the need for an empirical interpretation of any theory of confidence factors applied to Knowledge Based Systems and describes an empirical interpretation of DS theory suggesting that the theory has been extensively misinterpreted. For the essentially syntactic DS theory, a model is developed based on sample spaces, the traditional semantic model of probability theory. This model is used to show that, if belief functions are based on reasonably accurate sampling or observation of a sample space, then the beliefs and upper probabilities as computed according to DS theory cannot be interpreted as frequency ratios. Since many proposed applications of DS theory use belief functions in situations with statistically derived evidence [Wesley] and seem to appeal to statistical intuition to provide an interpretation of the results [Garvey], it may argued that DS theory has often been misapplied.

The success of the scientific approach is generally attributed by philosophers such as Popper to its insistence on empirical verification of theories [Davis]. Stated from a different point of view, theories which do not make empirically verifiable predictions about reality are not scientific. When one builds

Knowledge Based Systems for applications and includes the use of confidence factors, these confidence factors presumably are present to make some statement about the real world. If such Knowledge Based Systems are to be considered scientific, we must face the problem of empirically testing these statements. Arguments advanced in support of the various theories of confidence factors are almost never empirically testable. Some arguments often presented in favor of the DS theory are that the results are intuitively appealing, and that ranges of confidence are certainly better than point estimates [Shafer]. An argument often presented in favor the Fuzzy Sets is that humans are satisfied with the quantitative values produced, and that the theory "seems to work" [Zadeh '84].

The point of view taken here is that such arguments are sufficient to motivate further investigation into the merits of the these theories but are not sufficient to justify them as a basis for wholesale practical application, that ranges of confidence factors are useless if they have no empirically testable meaning, and that "seeming to work" arguments are subject to the Perceptron fallacy.

An attempt to provide an empirically testable semantic interpretation of DS is described in the remainder of this paper. The rather surprising result of this attempt is strong support for the conjecture that DS theory cannot, <u>in principle,</u> be empirically interpreted as calculus applicable to sample spaces, if the belief functions arise from the real world in a non-random way. Thus, though the numbers produced by this theory are often termed probabilites, they are not probabilites in the normal meaning employed in the experimental sciences. A connection seems likely between this conjecture and Zadeh's [Zadeh] observation that Dempster's evidence combination rule requires use of probability statements about undefined events.

The semantic model which provides the empirical basis for traditional probability models is the empirically defined sample space model, often discussed in terms of balls in an urn. Indeed, this model is usually considered an intrinsic part of traditional probability theory [Parzen] as developed by Fisher and von Mises. Certain aspects of traditional theory were axiomatized by Kolmolgorov but these axioms do not capture all aspects of traditional theory. In particular, **events** require no empirical definition in the Kolmolgorov theory. Theories such as DS and Fuzzy Sets, in general, satisfy the Kolmolgorov axioms but seem to lack the semantic interpretation possible with empirical sample space models.

A sample space interpretation of DS is now presented which provides a reasonable, meaningful (but of course not the only possible) empirical definition of the **belief function** of the DS theory. However, the application of Dempster's rule will be shown to destroy the sample space intepretation of combined belief functions. (We cannot, as yet argue that there is no possible sample space interpretation of DS whose meaning can be preserved by Dempster's rule, but it is our conjecture that under very non-restrictive conditions, that this is indeed the case.) The presentation begins with a brief, considerably simpler, interpretation of the Bayesian and Generalized Bayesian [Lemmer] inference in terms of sample spaces. This sample space model is then expanded to cover the DS theory.

As a convenient way of discussing sample spaces, consider an urn containing a number of balls, each of which is marked with zero or more different labels. For example some of the balls may be labeled "a", others "b", some may have both marks while others have neither. [These labels may be thought of as binary random variables: if label "a" is present, then variable "a" has value 1, otherwise it has value zero.] Assume that when observing the balls it is cheaper, in some sense, to check for the presence of some labels than others. [If we are restricted to "remote sensing" it will be cheaper to determine if the ball is greater than an inch in diameter than to determine if it weighs more than an ounce.] An experiment in such a model consists of drawing a single ball from the urn and observing some of its labels. The problem in classical probability theory is to estimate, before drawing the ball, the probability of drawing a ball labeled "a". This probability, under appropriate assumptions about how the ball is drawn, is usually taken to be the fraction of the balls in the urn which are labeled "a". This fraction can be determined by counting all the balls in the urn, or if this is infeasible, using the techniques of Sampling Theory to estimate this fraction.

The problem in Bayesian probability theory is to estimate, having drawn the ball and made an observation about label "a", the probability of the ball also being marked with label "b". Under Bayes' rule this probability is taken to be the fraction of balls having label "b" in that partition of the sample defined by balls having label "a". (Again, if the whole sample space is not available, techniques of Sampling Theory can be used to form estimates.)

The Generalized Bayesian problem addresses the case where one cannot be certain whether or not the label "a" is present. Instead, after drawing, one uses observation of the drawn ball to revise the previous estimate of the probability of the presence of label "a". The result of the observation, in the generalized case is that, with probability, p, the label "a" is present. (This probability itself could have been estimated based on previous samplings and similar observations.) Under the Generalized Bayes Rule, the probability of "b" on the drawn ball is again taken to be the fraction of balls with label "b" in a particular partition of the sample space. In this case however, the partition is formed from the previously discussed "a", "not a" partitions of the sample space. The new partition is constructed by randomly drawing balls from the old partitions in such a way that a fraction, p, come from the old "a" partition and the remainder from the old "not a" partition.

The major point to be observed here is that Sample Theory, Bayesian Theory, and Generalized Bayesian Theory are mathematical models of the empirical operations just described. The question to be addressed now is "of what empirical operations is Dempster's Rule the model?" In particular does it model similar operations on balls in an urn?

To give an empirical sample space interpretation for the belief functions of the DS theory, the balls in the urn must be labeled differently than just described. This labeling also highlights a difference in focus between the two theories: The Bayesian Theory is most suitable for balls which have multiple co-occurring labels and for observations which determine the presence or absence of a particular label type; the DS theory is most suitable for balls which have a single label drawn from a set of possible labels (the frame of discernment), and observations which restrict the label to some subset of the set of possible labels.

Imagine balls which have a single "true" label which, in some sense is hidden, and multiple other labels which are accessible to observation processes which we shall call sensors. The use of these sensors will be to form belief functions empirically. The
true label is drawn from the mutually exclusive and collectively
exhaustive set of labels usually termed the frame of discernment
or "theta". The labels accessible to the various sensors
correspond to subsets of theta. Thus the true label on a ball
might be "a" while the label accessible by some particular sensor
might be "a or b". The label available to some other belief
forming process might be "a or c". As an example of such a
labeling process, consider theta to be the set of all sailboats.
An observation of a particular sailboat in a photograph taken in
mist with an uncalibrated camera might reveal only that the
sailboat in question is a catamaran. This is equivalent to
labeling the observation with the subset of all sailboats,
{Hobie-Cat,Prindle,...}.

An experiment in this model consists of drawing a ball and
assigning a probability range to each possible proposition about
the true label of the ball. This assignment of ranges is done by
using Dempster's rule to combine a set of belief functions
concerning the balls in the urn and the particular ball which was
drawn. This assignment of ranges is accomplished by applying
Dempster's rule to a series of belief functions constructed about
the balls in the urn and the drawn ball. We will now define
empirical semantics for the belief functions to be combined.

For each sensor, develop a belief function from the DS urn in the following way: Let the probability mass, m(T), associated with each subset of theta, T, be the fraction of balls labeled T by the particular sensor. Then, if the sensor labeling is accurate, that is, if t*, the element of theta which is the "true" label of the particular ball, is always an element of the label applied to that ball, then the minimum fraction of balls for which t* is an element of T is given by

$$B(T) = \sum_{t \subset T} m(t) \qquad (1)$$

which is in accord with the DS definition of Bel(T). The maximum fraction of balls which could have a true label which is an element of T is given by

$$p^*(T) = \sum_{\substack{U \subset \text{theta} \\ U \cap T \neq \{\}}} m(U) = B(T) + \sum_{\substack{U \subset \text{theta} \\ U \cap (\text{theta} - T) \neq \{\}}} m(U) \qquad (2)$$

which is in accord with the DS definition of p*(T). Equations (1) and (2) follow from the assumption that every sensor labeling is accurate. If the sensor labelings are not accurate, we have no assurance that (1) and (2) define the minimum and maximum fractions. In fact if the labelings can be false, we can say nothing about the possible range of the fractions. Thus, under the assumption of accurate sensor labeling, we have an empirical definition of individual belief functions corresponding to the DS definition.

It is generally stated that if belief functions are independent, then Dempster's rule, shown in (3) below, is a valid way to combine individual belief functions. If the combined belief function is to be empirically meaningful, then equations (1) and (2) using the probability mass from the combined belief function should provide ranges which are true with respect to the balls in the urn.

We will now show that there are labeling processes, independent of each other in every practical sense (to be defined), which give rise to belief functions which when combined, do not provide ranges accurately describing the original population. These labeling processes will be seen to be of great practical importance, not just of academic interest. We will, however, show the converse that given two belief functions, a population can always be generated which satisfies the combined belief function. We will show the former by example, and the latter by providing a construction. The construction provides insight as to why population parameters cannot be described by combined belief functions developed from accurate sensors and provides a connection to the claims of Zadeh mentioned above.

Consider an urn containing balls whose true labels are exactly one of the labels from the the set {a,b,c}. To provide intuition concerning the action of sensors, assume that balls which are labeled "a" are light in weight while balls labeled either "b" or "c" are heavy. Assume also that balls labeled either "a" or "b" are red in color while balls labeled "c" are blue. Thus two sensors, one of which can classify weight and the other which can classify color can each be thought of as a process for empirically determining belief functions. Sensor 1 can be thought of as labeling balls, truly labeled, "a" with a label $t1=a$, and balls, truly labeled "b" or "c", with a label $t1=b+c$ (read "b" or "c"). Likewise sensor 2 provides either $t2=a+b$ or $t2=c$. If $p(a)$, $p(b)$, and $p(c)$ represent the actual fraction of balls having true labels "a", "b", and "c" respectively, then the two sensor systems could be used to empirically determine the belief functions illustrated in Table 1.

| Sensor 1 | Sensor 2 |
|---|---|
| $m1(a) = p(a)$ | $m2(a+b) = p(a+b)$ |
|  | $= p(a) + p(b)$ |
| $m1(b+c) = p(b+c)$ |  |
| $= p(b) + p(c)$ | $m2(c) = p(c)$ |

Table 1: Two Belief Functions

It can readily be seen that the labeling processes are independent when conditioned by the value of the true label but are not unconditionally independent. This is exactly as would be desired in real applications since unconditional independence would imply a non-zero probability of inconsistent labeling by the two sensors, implying one or both sensors applied labels of which the true label was not an element. In the real world one presumably always strives to have error free sensors, and would certainly hope to have sensors whose actual error rate is less than that required to produce unconditionally independent labeling.

We now examine the belief function obtained by using Dempster's rule, (3), to combine the two belief functions shown in Table 1.

$$m3(T) = \frac{1}{k} \left[ \sum_{T \subset U \subset \theta} [m1(T) \, m2(U) + m1(U) \, m2(T)] - m1(T) \, m2(T) \right]$$

where  (3)

$$k = 1 - \sum_{\substack{x \subset \theta \; y \subset \theta \\ x \cap y = \{\}}} m1(x) \, m2(y)$$

Applying (3) to our example where theta ={a,b,c} yields

$$p(a) \Leftarrow B(a) = m3(a) = \frac{m1(a)\, m2(a+b)}{1 - m1(a)m2(c)} \qquad (4)$$

where the inequality expresses the claim we are investigating. The first equality is from (2) and the second equality is from (3). Substituting the empirical belief values from Table 1 gives

$$p(a) \Leftarrow \frac{p(a)\, p(a+b)}{1 - p(a)p(c)} \qquad (5)$$

Since a and b are mutually exclusive events, we may substitute p(a)+p(b) for p(a+b). The elements of the frame of discernment are mutually exclusive and collectively exhaustive so we may substitute 1-p(c) for p(a)+p(b). Making these substitutions and dividing through by p(a) produces

$$1 \Leftarrow \frac{1 - p(c)}{1 - p(a)p(c)} \qquad (6)$$

Since p(a), p(b), and p(c) must all be positive and recalling that we divided through by p(a), we can see that that (6) holds only if p(c) is 0. If it is not, then the belief function resulting from the application of Dempster's rule does not correctly bound the value of p(a) in the population. Note that if p(c) is 0, then the labeling processes which produced the belief functions in Table 1

172

are unconditionally as well as conditionally independent.

The first major point is that accurate labeling processes are never unconditionally independent except in trivial cases. By an accurate labeling process we mean a process which does not assign a label which is inconsistent with the true label. In general, such inconsistent labelings must occur if the labeling processes are unconditionally independent. Indeed the normalizing factor in equation (3) can, in our semantic model, be interpreted as one minus the probability of inconsistent labeling, assuming unconditionally independent labeling processes. We can conclude that DS is not appropriate for dealing with sensors which provide accurate labelings.

The second major point is that, in our empirical interpretation, the estimation of a belief function is no different, in principle, than the estimation of a Bayesian prior. Both are obtained either by examining the entire population or by using Sampling Theory (or similar approaches) to estimate parameters of the entire population.

We now turn our attention to _constructing_ a population which is consistent with Dempster's combination of belief functions. Before, we imagined that the balls in the urn had true labels initially and were further labeled by sensors which had access to some information about the true labels. Now we assume that the balls have no true labeling at the beginning and that a true
labeling will be generated only after the combined belief function
is determined. We apply our sensor labeling process as follows:
Allow sensor 1 to randomly label each ball but in such a way that
the total fraction of balls receiving a label, T1=x, is equal to
m1(x). (There is no restriction on Bell other than that the m1(x)
be non-negative and sum to one over all subsets of theta.) Allow
sensor 2 to label the balls in a similar random fashion, taking
special care that second labeling is independent of the first (i.e
p({T2=y}/{T1=x})=p({T2=y})=m2(y)). Now examine each ball and
assign it a true label from the set x intersect y. If the
intersection is empty discard the ball. Note that discarding these
balls is the semantic equivalent of the normalization operation in
equation (3). The result will be a population of balls such that
equations (2) and (3) accurately describe the range of true
probabilities which might occur in the population. The actual
probabilities of course are determined by how the true label is
chosen from the intersection of x and y. The minimum probability
is achieved for p(t) if t is never chosen as the true label unless
x .int. y = {t} and the maximum for p(t) is achieved if t is
always chosen as the true label if it is an element of x.int.y.
Note, however that the ranges are not independent. For example,
no population exists in which all minima are achieved
simultaneously.

The two labeling methods can be viewed in the following way. In the first labeling method, that in which a pre-existing population is examined, the sensors had access to some partial knowledge concerning the true label of the balls. These true labels are conceived of as existing before and independently of the sensing process. In the second method, that of constructing a population, the sensors acted independently of each other and with no knowledge of the true label. True labels can be assigned only after the final belief is formed. Thus in dealing with a sample space we are led to a paradox: If we accurately observe the population, we cannot use Dempster's rule; if we use Dempster's rule we must form our beliefs without studying the population.

Of course there may exist other semantic models of the Dempster-Shafer theory which avoid the problems and paradox described above. But until such models are developed, it is our conclusion that Dempster's rule is not applicable to situations which can be modeled by sample spaces in which the sample points have "true" labels independent of the sensing process. While the DS method produces ranges of belief rather than point estimates, these ranges will not be related to frequencies of occurrence in meaningful situations. Thus these ranges cannot be used as input to decision theoretic processes and cannot be used, for example, to estimate expected utility. It appears to be an open question as to how decisions should be made, based on the outputs of DS calculations, in any situation in which frequencies of occurrence, predicted error rates, etc. are of concern.